\documentclass[a4paper]{article}

\usepackage{graphicx}
\usepackage{subfigure} 
\usepackage{fancyhdr}
\usepackage{fancybox}
\usepackage{natbib}
\usepackage{hyperref}
\usepackage{xcolor}
\usepackage{amsmath,amssymb}

\usepackage{hyperref}

\usepackage{float}
\newtheorem{hyp}{Hypothesis}

\newtheorem{proposition}{Proposition}[section]
\newtheorem{definition}[proposition]{Definition}

\newtheorem{theorem}{Theorem}
\newcommand{\bpf} {\noindent{\sc Proof} : }
\newcommand{\epf} {\hfill$\square$\vspace{.5cm}}

\newcommand{\f} {\mathbf{f}}
\renewcommand{\P} {\mathbb{P}}

\newcommand{\N}{\mathbb{N}}
\newcommand{\R}{\mathbb{R}}

\newcommand{\h}{\mathcal{H}}

\newcommand {\ep} {\varepsilon}
\newcommand {\x} {\mathcal{X}}
\newcommand {\y} {\mathcal{Y}}

\renewcommand {\l} {\ell}

\DeclareMathOperator*{\argmin}{arg\,min} 
\usepackage{color}
\definecolor{MyDarkBlue}{rgb}{0.05,0.,0.8}

\begin{document} 

\lhead{J. Audiffren, H. Kadri} 
\rhead{Online Learning with Multiple Operator-valued Kernels}
\rfoot[Technical Report V 1.0]{\thepage} 
\cfoot{} 
\lfoot[\thepage]{Technical Report V 1.0}

\renewcommand{\headrulewidth}{0.4pt}  
\renewcommand{\footrulewidth}{0.4pt}

\title{Online Learning with Multiple Operator-valued Kernels}

 \author{ Julien Audiffren 
\and Hachem Kadri
\and  Aix-Marseille Univ., LIF-QARMA, CNRS, UMR 7279, F-13013, Marseille, France\\ \textit{\{firstname.name\}@lif.univ-mrs.fr}}

\maketitle

\begin{abstract}

We consider the problem of learning a vector-valued function $f$ in an online learning setting. The function $f$ is assumed to lie in a reproducing Hilbert space of operator-valued kernels. We describe two online algorithms for learning $f$ while taking into account the output structure. A first contribution is an algorithm, ONORMA, that extends the standard  kernel-based online learning algorithm NORMA from scalar-valued to operator-valued setting.  We report a cumulative error bound that holds both for classification and regression. We then define a second algorithm, MONORMA, which addresses the limitation of pre-defining the output structure in ONORMA by learning sequentially a linear combination of operator-valued kernels. Our experiments show that the proposed algorithms achieve good performance results with low computational cost.
\end{abstract}

\section{Introduction}



We consider the problem of learning a function $f:\x \rightarrow \y$ in a reproducing kernel Hilbert space, where $\y$ is a Hilbert space with dimension $d>1$. This  problem has received relatively little attention in the machine learning community compared to the analogous scalar-valued case where $\y \subseteq R$. In the last decade, more attention has been paid on learning vector-valued functions~\citep{micchelli05}. This attention is due largely to the developing of practical machine learning~(ML) systems, causing the emergence of new ML paradigms, such as multi-task learning and structured output prediction, that can be suitably formulated as an optimization of vector-valued functions~\citep{Evgeniou05, kadri13}.

Motivated by the success of kernel methods in learning scalar-valued functions, in this paper we focus our attention to vector-valued function learning using reproducing kernels (see~\citealp{Alvarez12}, for more details). It is important to point out that in this context the kernel function outputs an operator rather than a scalar as usual. The operator allows to encode prior information about the outputs, and then take into account the output structure. Operator-valued kernels have been applied with success in the context of multi-task learning~\citep{Evgeniou05}, functional response regression~\citep{kadri10} and structured output prediction~\citep{brouard11, kadri13}. Despite these recent advances, one major limitation with using operator-valued kernels is the high computational expense. Indeed, in contrast to the scalar-valued case,  the kernel matrix associated to a reproducing operator-valued kernel is a block matrix of dimension $td \times td$, where $t$ is the number of examples and $d$ the dimension of the output space. Manipulating and inverting matrices of this size become particularly problematic when dealing with large $t$ and $d$. In this spirit we have asked whether, by learning the vector-valued function $f$ in an online setting, one could develop efficient operator-valued kernel based algorithms with modest memory requirements and low computational cost.

Online learning has been extensively studied in the context of kernel machines~\citep{Smola04, Crammer2006,  Cheng06, Ying06, Vovk06, Orabona09, Cesa2010, Zhang13}. We refer the reader to~\citet{Girolami13} for a review of kernel-based online learning methods and a more complete list of relevant references. However,  most of this work has focused on scalar-valued kernels and little attention has been paid to operator-valued kernels. The aim of this paper is to propose {new} algorithms for online learning with operator-valued kernels suitable for multi-output learning situations and able to make better use of complex (vector-valued) output data to improve non-scalar predictions.
It is worth mentioning that recent studies have adapted existing batch multi-task methods to perform online multi-task learning~\citep{Cavallanti:2010, Dinuzzo10, Saha2011}. Compared to this literature, our main contributions in this paper are as follows:
\begin{itemize}
\item we propose a first algorithm called ONORMA, which extends the widely known NORMA algorithm~\citep{Smola04} to operator-valued kernel setting,

\item we provide theoretical guarantees of our algorithm that hold for both classification and regression,

\item we also introduce  a second algorithm, MONORMA, which addresses the limitation of pre-defining the output structure in ONORMA by learning sequentially both a vector-valued function and a linear combination of operator-valued kernels,

\item we provide an empirical evaluation of the performance of our algorithms which demonstrates their effectiveness on synthetic and real-world multi-task data sets.

\end{itemize}

%
%

\section{Overview of the problem}

This section presents the notation we will use throughout the paper and introduces the problem of online learning with operator-valued kernels.

\textbf{Notation.}
Let $(\Omega, \mathcal{F},\P)$ be a probability space, $\x$ a Polish space, $\y$ a (possibly infinite-dimensional) separable Hilbert space, $\h$ a separable Reproducing Kernel Hilbert Space~(RKHS) $\subset \y^\x$ with $K$ its Hermitian positive-definite reproducing operator-valued kernel, and $L(\y)$ the space of continuous endomorphisms\footnote{We denote by $My=M(y)$ the application of the operator $M \in L(\y)$ to $y\in\y$.}  of $\y$ equipped with the operator norm. Let $t$ $\in \mathbb{N}$ denotes the number of examples, $(x_i,y_i) \in \mathcal{X}\times\mathcal{Y}$ the $i$-th example, $\l:\y \times \y \to \R^+$ a loss function, $\nabla$  the gradient operator, and $R_{inst}(f,x,y)= \l(f(x),y)+(\lambda/2) \|f\|_\h^2$ the instantaneous regularized error. Finally let $\lambda \in \R$ denotes the regularization parameter and $\eta_t$ the learning rate at time $t$, with $\eta_t \lambda <1$.

We now briefly recall the definition of the operator-valued kernel $K$ associated to a RKHS $\h$ of functions that maps $\x$ to $\y$. For more details see~\citet{micchelli05}.
\begin{definition}\label{noyau HS}
The application $K : \x \times \x \rightarrow L(\y) $ is called the Hermitian positive-definite reproducing operator-valued kernel of the RKHS $\h$ if and only if :
\begin{enumerate}
\item $\forall x \in \x,\forall y \in \y,$ the application 
\begin{equation}\nonumber
\begin{aligned}
 K(.,x)y : \x &\to \y \\
 			x' &\mapsto K(x',x)y
\end{aligned}
\end{equation} 
belongs to $\h$.

\item $\forall f \in \h$, $\forall x \in \x,$ $\forall y \in \y,$
$$ \left\langle f(x),y \right \rangle_\y = \left\langle f,K(.,x)y \right \rangle_\h, $$
\item $\forall x_1,x_2\! \in \x,$ 
\\[0.2cm] $~ \qquad$ $ K(x_1,x_2)\!=\!K(x_2,x_1)^*\! \in\! L(\y)$ ($*$ denotes the adjoint), 
\item $\forall n \ge 1,$ $\forall (x_i,i\in \{1..n\}), (x'_i,i\in \{1..n\}) \in \x^n$, $\forall (y_i,i\in \{1..n\}), (y'_i,i\in \{1..n\}) \in \y^n$,
$$ \sum_{k,\l=0}^n \left\langle K(.,x_k)y_k,K(.,x'_\l)y'_\l \right \rangle_\y  \ge 0.$$
\end{enumerate}
(i) and (ii) define a reproducing kernel, (iii) and (iv) corresponds to the Hermitian and  positive-definiteness properties, respectively.
\end{definition}

\noindent\textbf{Online learning.} In this paper we are interested in online learning in the framework of operator-valued kernels. On the contrary to batch learning, examples $(x_i,y_i)\in \x \times \y$ become available sequentially, one by one. The purpose is to construct a sequence $(f_i)_{i \in \N}$ of elements of $\h$ such that:
\begin{enumerate}
\item $\forall i \in \N$, $f_i$ only depends on the previously observed examples $\left\lbrace (x_j,y_j), 1 \le j \le  i \right\rbrace$,
\item The sequence  $(f_i)_{i \in \N}$ try to minimize $ \sum_i R_ {inst}(f_i,x_{i+1},y_{i+1})$.
\end{enumerate}

In other words, the sequence of $f_t$, $t\in \{1,\ldots,m\}$, tries to predict the relation between $x_t$ and $y_t$ based on previously observed examples. 

In section \ref{sect ONORMA} we consider the case where the functions $f_i$ are constructed in a RKHS $\h \subset \y^\x$. We propose an algorithm, called ONORMA, which is an extension of NORMA \citep{Smola04} to operator-valued kernel setting, 
and we prove theoretical guarantees that holds for both classification and regression.

In section \ref{sect OMKL} we consider the case  where the functions $f_i$ are constructed as a linear combination of functions $g_i^j$, $1 \le j \le m$, of different RKHS $\h^j$. We propose an algorithm which construct both the $j$ sequences $(g_i^j)_{i\in \N}$ and the coefficient of the linear combination. This algorithm, called MONORMA,  is a variant of ONORMA that allows to learn sequentially the vector-valued function $f$ and a linear combination of operator-valued kernels. MONORMA can be seen as an online version of the recent  MovKL (multiple operator-valued kernel  learning) algorithm \citep{kadri12}.

It is important to note that both algorithms, ONORMA and MONORMA,  do not require to invert the block kernel matrix associated to the operator-valued kernel and have at most a linear complexity with the number of examples at each update. This is a really interesting property 
 since in the operator valued kernel framework, the complexity of batch algorithms is extremely high and the inversion of kernel block matrices with elements in $L(\y)$ is an important limitation in practical deployment of  such methods. For more details, see `Complexity Analysis' in Sections~\ref{sect ONORMA} and~\ref {sect OMKL}.


\section{ONORMA}\label{sect ONORMA}

In this section we consider online learning of a function $f$ in the RKHS $\h \subset \y^\x$ associated to the operator-valued kernel $K$.  The key idea here is to perform a stochastic gradient descent with respect to $R_{inst}$ as in \citet{Smola04}. The update rule (i.e., the definition of $f_{t+1}$ as a function of $f_t$ and the input $(x_{t},y_{t})$) is
\begin{equation}\label{eq f_t 1}
 f_{t+1}= f_{t} - \eta_{t+1} \nabla_{f} R_{inst}(f,x_{t+1},y_{t+1})\vert_{f=f_t},
\end{equation}
where $\eta_{t+1}$ denotes the learning rate at time $t+1$.
Since $\l(f(x),y)$ can be seen as the composition of the following functions:
\begin{equation}\nonumber
\begin{aligned}
\h  \to \y  & \to \R\\
f \mapsto f(x) & \mapsto \l(f(x),y),\\
\end{aligned}
\end{equation}
 we have, $\forall h \in\h$,
\begin{equation}\nonumber
\begin{aligned}
D_f   \l(f(x)&,  y) \vert_{f=f_t}(h) \\ &=D_z \l(z,y)\vert_{z=f_t(x)}(D_f f(x)\vert_{f=f_t} (h) ) \\
&=D_z \l(z,y)\vert_{z=f_t(x)}(h(x))\\
&= \left\langle \nabla_{z} \l(z,y)\vert_{z=f_t(x)},h(x) \right\rangle_\y \\
&=\left\langle K(x,\cdot) \nabla_{z} \l(z,y)\vert_{z=f_t(x)},h \right\rangle_\h,
\end{aligned}
\end{equation}
where we used successively the linearity of the evaluation function, the Riesz representation theorem and the reproducing property. From this we deduce that
\begin{equation}\label{eq gradient}
\nabla_{f} (\l(f(x),y)\vert_{f=f_t}=K(x,\cdot) \nabla_{z} \l(z,y)\vert_{z=f_t(x)}.
\end{equation}
Combining \eqref{eq f_t 1} and \eqref{eq gradient}, we obtain
\begin{equation}\label{eq f_t 2}
\small
\begin{aligned}
 f_{t+1}&= f_{t} - \eta_{t+1}  \left( K(x_{t+1},\cdot) \nabla_{z} \l(z,y_{t+1})\vert_{z=f_t(x_{t+1})}+\lambda f \right)\\
 &= \left( 1 - \eta_{t+1} \lambda \right) f_{t} - \eta_t K(x_{t+1},\cdot) \nabla_{z} \l(z,y)\vert_{z=f_t(x_{t+1})}.
\end{aligned}
\end{equation}
From the equation above we deduce that if we choose $f_0=0$, then 
 there exists $\left( \alpha_{i,j} \right)_{i,j} $ a family of elements of $\y$  such that $\forall t\ge 0$,  
\begin{equation}\label{eq f_t 3}
\begin{aligned}
f_t= \sum_{i=1}^t K(x_i,\cdot)\alpha_{i,t} ,
\end{aligned}
\end{equation}
Moreover, using \eqref{eq f_t 2} we can easily compute the $\left( \alpha_{i,j} \right)_{i,j}$ iteratively. This is Algorithm 1, called ONORMA, which is an extension of NORMA \citep{Smola04} to operator-valued setting.

Note that \eqref{eq f_t 3} is a similar expression to the one obtained in batch learning framework, where the representer theorem~\cite{micchelli05} implies that the solution $f$ of 
 a regularized empirical risk minimization problem in the vector-valued function learning setting can be written as $f= \sum K(x_i,\cdot) \alpha_i$ with $\alpha_i \in \y$.

  \begin{table}
\begin{center}
  \begin{tabular}{l}
    \hline \\[-1.0em]
    \textbf{Algorithm 1} \ \ ONORMA\\
    \hline\\[-0.5em]
    \textbf{Input:}  parameter $\lambda,\eta_t\in \mathbb{R^*_+}$, $s_t \in \N$, \\ $\ \quad\qquad$ loss function $\l$, \\[0.1cm]
     \textbf{Initialization:}  $f_0$=0\\[0.1cm]
     \textbf{At time t:}  Input : ($x_t$,$y_t$) \\[0.1cm]

    $\quad$ 1.\  \  \textbf{New coefficient:}   \\[0.1cm]
   \hspace{1.8cm} $\alpha_{t,t} := - \eta_t \nabla_z \l(z,y_t)\vert_{z=f_{t-1}(x_t)}$  \\[0.2cm]
    $\quad$ 2.\  \  \textbf{Update old coefficient:} $\forall 1\le i\le t-1,$\\[0.1cm]
   \hspace{1.8cm} $\alpha_{i,t} := (1- \eta_t \lambda) \alpha_{i,t-1}$  \\[0.2cm]
    $\quad$ 3.\  \  \textbf{(Optional) Truncation:} $\forall 1\le i\le t-s_t,$ \\[0.1cm]
   \hspace{1.8cm} $\alpha_{i,t} := 0$  \\[0.2cm]
    $\quad$ 4.\  \  \textbf{Obtain $f_t$:}\\[0.1cm]
   \hspace{1.8cm} $f_t=\sum_{i=1}^{i=t} K(x_i,\cdot) \alpha_{i,t}$ \\[0.2cm] 
    \hline
  \end{tabular}
    \end{center}
  \end{table}

\textbf{Truncation.} Similarly to the scalar-valued case, ONORMA can be modified by adding a truncation step. In its unmodified version, the algorithm needs to keep in memory all the previous input $\{x_i\}_{i=1}^t$ to compute the prediction value $y_t$, which can be costly. However, the influence of these inputs decreases geometrically at each iteration, since $0< 1- \lambda \eta_t < 1$. Hence the error induced by neglecting old terms can be controlled  (see Theorem \ref{th bound cumulative loss truncate} for more details).  The optional truncation step in Algorithm 1 reflects this possibility. 

\textbf{Cumulative Error Bound.} The cumulative error of the sequence $(f_i)_{i \le t}$, is defined by $\sum_{i=1}^t \l(f_i(x_i),y_i)$, that is to say the sum of all the errors made by the online algorithm when trying to predict a new output.
It is interesting to compare this quantity to the error made with the function $\f_t$ obtained from a regularized empirical risk minimization defined as follows
%
\begin{equation}\nonumber
\begin{aligned}
\f_t &= \argmin_{h\in\h} R_{reg}(h,t)\\
&= \argmin_{h\in\h} \frac{1}{t}\sum_{i=1}^t \l(h(x_i),y_i) + \frac{\lambda}{2} \|h\|^2_\h .
\end{aligned}
\end{equation}
%
%
%
To do so, we make the following assumptions:
\begin{hyp}\label{hyp k bounded}
$\sup_{x \in \x} \vert k(x,x)\vert_{op} \le \kappa^2$.
\end{hyp}
\begin{hyp}\label{hyp sigma admissible}
$\l$ is $\sigma$-admissible,i.e. $\l$ is convex and $\sigma$-Lipschitz with regard to $z$.
\end{hyp}
\begin{hyp}\label{hyp least square}
\item
\renewcommand{\theenumi}{\roman{enumi}}%
\begin{enumerate}
\item $\l(z,y)= \frac{1}{2}\| f(x) - z \|^2_\y $ ,
\item  $\exists C_y>0$  such that $\forall t \ge 0$, $\|y_t\|_\y \le C_y$,
\item $\lambda > 2 \kappa^2$.
\end{enumerate}
\end{hyp}

The first hypothesis is a classical assumption on the boundedness of the kernel function.
Hypothesis \ref{hyp sigma admissible} requires the $\sigma$-admissibility of the loss function $\l$, which is also a common assumption.
Hypothesis~\ref{hyp least square} provides a sufficient condition to recover the cumulative error bound  in the case of least square loss function, which is  not $\sigma$-admissible.
It is important to note that while Hypothesis \ref{hyp sigma admissible} is a usual assumption in the scalar case, bounding  the least square cumulative error in regression using Hypothesis \ref{hyp least square}, as far as we know, is a new approach.


\begin{theorem}\label{th bound cumulative loss}
Let $\eta >0$ such that $\eta \lambda <1$. If Hypothesis \ref{hyp k bounded} and either Hypothesis \ref{hyp sigma admissible} or \ref{hyp least square} holds, then there exists $U>0$ such that, with $\eta_t=\eta t^{-1/2}$,
$$\frac{1}{m} \sum_{i=1}^m R_{inst}(f_i,x_i,y_i) \le  R_{reg}(\f_m,m) +\frac{\alpha}{\sqrt{m}}+ \frac{\beta}{m},$$
where $\alpha=2 \lambda U^2(2\eta \lambda + 1/(\eta \lambda))$, $\beta= U^2/(2 \eta)$, and $f_t$ denotes the function obtained at time $t$ by Algorithm~1 without truncation.
\end{theorem}

A similar result can be obtained in the truncation case.

\begin{theorem}\label{th bound cumulative loss truncate}
Let $\eta >0$ such that $\eta \lambda <1$ and $0< \ep <1/2$. Then, $\exists t_0>0$ such that for $$s_t = \min(t,t_0) + 1_{t>t_0} \lfloor (t-t_0)^{1/2+\ep} \rfloor, $$  if Hypothesis \ref{hyp k bounded} and either Hypothesis \ref{hyp sigma admissible} or \ref{hyp least square} holds, then there exists $U>0$ such that, with $\eta_t=\eta t^{-1/2}$,
$$\frac{1}{m} \sum_{i=1}^m R_{inst}(f_i,x_i,y_i) \le \inf_{g \in \h} R_{reg}(\f_m,m) +\frac{\alpha}{\sqrt{m}}+ \frac{\beta}{m},$$
where $\alpha=2 \lambda U^2(10\eta \lambda + 1/(\eta \lambda))$, $\beta= U^2/(2 \eta)$ and  $f_t$ denotes the function obtained at time $t$ by Algorithm~1 with truncation.
\end{theorem}

The proof of Theorems 1 and 2 differs from the scalar case through several points which are grouped in Propositions \ref{prop admissible} (for the $\sigma$-admissible case) and \ref{prop LSR} (for the least square case). Due to the lack of space, we present here only the proof of Propositions \ref{prop admissible} and \ref{prop LSR}, and refer the reader to \citet{Smola04} for the proof of Theorems 1 and 2.

\begin{proposition}\label{prop admissible}
If Hypotheses \ref{hyp k bounded} and \ref{hyp sigma admissible} hold, we have
\renewcommand{\theenumi}{\roman{enumi}}%
\begin{enumerate}
\vspace{-0.1cm}
\item $\forall t \in \N^*,$ $\|\alpha_{t,t}\|_\y \le \eta_t C$,
\item if $\| f_0 \|_\h \le U=C\kappa / \lambda$ then $\forall t \in \N^*,$ $\| f_t \|_\h \le U$,
\item$\forall t \in \N^*, \|\f_t \|_\h \le U$.
\end{enumerate}
\end{proposition}

\bpf$(i)$ is a consequence of the Lipschitz property and the definition of $\alpha$,
\begin{equation}\nonumber
\begin{aligned}
\| \alpha_{t,t} \|_\y= \| \eta_t \nabla_z \l(z,y_t)\vert_{z=f_{t-1}(x_t)} \|_\y \le \eta_t C.
\end{aligned}
\end{equation}
$(ii)$ is proved using Eq.~\eqref{eq f_t 2} and $(i)$ by induction on $t$. (ii) is true for $t=0$. If (ii) is true for $t=m$, then is is true for $m+1$, since
\begin{equation}\nonumber
\begin{aligned}
 \|f_{m+1}\|_\h &= \| \left( 1 - \eta_t \lambda \right) f_{m} +  k(x,\cdot) \alpha_{m,m} \|_\h\\
 &\le  \left( 1 - \eta_t \lambda \right) \| f_{m} \|_\h + \sqrt{\|k(x,x)\|_{op}} \| \alpha_{m,m} \|_\h\\
 &\le \frac{C\kappa}{\lambda} -\kappa \eta_t C +  \kappa \eta_t C = \frac{C\kappa}{\lambda}.
\end{aligned}
\end{equation}
To prove $(iii)$, one can remark that by definition of $\f_t$, we have $\forall \ep >0$,
\begin{equation}\nonumber
\begin{aligned}
0 &\le \frac{\lambda}{2} (\|(1-\ep) \f\|^2_\h - \|\f\|^2_\h ) \\
&\hspace{1cm}+ \frac{1}{t}\sum_{i=1}^t \l ( (1-\ep)\f(x_i),y_i) - \l ( \f(x_i),y_i) \\
&\le \frac{\lambda}{2} ( \ep^2 - 2 \ep) \| \f \|^ 2_\h + C \ep \kappa \|\f\|_\h.
\end{aligned}
\end{equation}
Since this quantity must be positive for any $\ep>0$, the dominant term in the limit when $\ep\to 0$, i.e. the coefficient of $\ep$,  must be positive.
Hence $\lambda  \| \f \|_\h \le C \kappa$. \epf

\vspace{-0.4cm}

\begin{proposition}\label{prop LSR}
If Hypotheses \ref{hyp k bounded} and \ref{hyp least square} hold, and  $\|f_0\|_\h < C_y/\kappa \le U=\max(C_y /\kappa,2C_y / \lambda )$, then $\forall t \in \N^*$:
\renewcommand{\theenumi}{\roman{enumi}}%
\begin{enumerate}
\vspace{-0.1cm}
\item $ \|f_t\|_\h < C_y/\kappa \le U$,
\item $\exists V$  in a `neighbourhood' of $f_t$ such that  $\l(\cdot,y_t)\vert_V$  is $2C_y$ Lipschitz,
\item  $\|\alpha_{t,t}\|_\y \le 2\eta_t C_y$,
\item  $ \|\f_t\|_\h \le \frac{2C_y}{\lambda} \le U$.
\end{enumerate}
\end{proposition}
\bpf We will first prove that $\forall t \in \N^* $, 
$$(i)\implies(ii) \implies (iii).$$
\textbf{$(i)\implies(ii)$:} In the Least Square case, the application $z \mapsto \nabla_z \l(z,y_t) =  (z-y_t) $ is continuous. Hypothesis \ref{hyp least square}-$(ii)$ and \ref{hyp k bounded} combined with $(i)$ imply that  $\| \nabla_z \l(z,y_t)\vert_{z=f_{t}(x_t)} \|_\y < 2C_y$. Using the continuity property,  we obtain $(ii)$.

\noindent\textbf{$(ii)\implies(iii)$:} We use the same idea as in Proposition~\ref{prop admissible}, and we obtain $$\| \alpha_{t,t} \|_\y= \| \eta_t \nabla_z \l(z,y_t)\vert_{z=f_{t-1}(x_t)} \|_\y \le \eta_t 2C_y .$$

Now we will prove $(i)$ by induction:

\noindent\textit{Initialization (t=0).} By hypothesis, $\|f_0\|_\h < C_y/\kappa$.

\noindent\textit{Propagation (t=m):} If $\|f_m\|_\h < C_y/\kappa$, then using $(ii)$ and $(iii)$:
\begin{equation}\nonumber
\begin{aligned}
 \|f_{m+1}\|_\h &\le  \left( 1 - \eta_t \lambda \right) \| f_{m} \|_\h + \sqrt{\|k(x,x)\|_{op}} \| \alpha_{m,m} \|_\h\\
 &\le \left( 1 - \eta_t \lambda \right) C_y/\kappa + 2 \kappa \eta_t C_y \\
 &= C_y/\kappa + \eta_t C_y ( 2\kappa -\lambda / \kappa  )\\
 &< C_y/\kappa,
\end{aligned}
\end{equation}
where the last transition is a consequence of Hypothesis \ref{hyp least square}-$(iii)$.

Finally, note that by definition of $\f$, since $0\in\h$,
\begin{equation}\nonumber
\begin{aligned}
\frac{\lambda}{2}  \|\f\|^2_\h + \frac{1}{t}\sum_{i=1}^t \l ( \f(x_i),y_i)  &\le \frac{\lambda}{2}  \|0\|^2_\h + \frac{1}{t}\sum_{i=1}^t \l ( 0,y_i) \\
&\le C_y^2.
\end{aligned}
\end{equation}
Hence,  $\|\f\|_\h \le 2C_y/\lambda$. \epf

\textbf{Complexity analysis.}
Here we consider a naive implementation of ONORMA algorithm when dim$\y =d < \infty$. 
At iteration $t$, the calculation of the prediction has complexity $O(t d^2)$ and the update of the old coefficient has complexity $O(t d)$.
In the truncation version, the calculation of the prediction has complexity $O(s_n d^2)$ and the update of the old coefficient has complexity $O(s_t d)$. The truncation step has complexity $O(s_t-s_{t-1})$ which is $O(1)$ since $s_t$ is sublinear.
Hence the complexity of the iteration $t$ of \mbox{ONORMA} is $O(s_t d^2)$ with truncation (respectively $O(t d^2)$ without truncation) and the complexity up to iteration $t$ is respectively $O(t s_t d^2)$ and $O(t^2 d^2)$. Note that the complexity of a naive implementation of the batch algorithm for learning a vector-valued function with operator-valued kernels is $O(t^3d^3)$. A major advantage of ONORMA is  its lower computational complexity compared to classical batch operator-valued kernel-based algorithms.

\section{MONORMA}\label{sect OMKL}

\begin{table*}[!t]
    \begin{center}
  \begin{tabular}{l}
    \hline \\[-1.0em]
    \textbf{Algorithm 2}   \ \ MONORMA  \hspace{5.9cm} \\
    \hline\\[-0.5em]
    \textbf{Input:}  parameter $\lambda,r>0$,$\eta_t\in \mathbb{R^*_+}$, $s_t \in \N$, loss function $\l$, kernels $K^1$,...,$K^m$ \\[0.1cm]
     \textbf{Initialization:}  $f_0=0$,$g^j_0=0$, $\delta^j_0= \frac{1}{n}$,$\gamma^j_0=0$\\[0.1cm]
     \textbf{At time $t\ge1$:}  Input : ($x_t$,$y_t$) \\[0.1cm]
    $\quad$ 1.\  \ \textbf{$\alpha$ update phase:}\\[0.1cm]
    $\quad$ $\quad$ $\quad$ 1.1	\  \  \textbf{New coefficient:}   \\[0.1cm]
   \hspace{4cm} $\alpha_{t,t} := - \eta_t \nabla_z \l(z,y_t)\vert_{z=f_{t-1}(x_t)}$  \\[0.2cm]
    $\quad$ $\quad$ $\quad$ 1.2\  \  \textbf{Update old coefficient:} $\forall 1\le i\le t-1,$\\[0.1cm]
   \hspace{4cm} $\alpha_{i,t} := (1- \eta_t \lambda) \alpha_{i,t-1}$  \\[0.2cm]
   $\quad$ 2. \textbf{$\delta$ update phase:}\\[0.1cm]
   $\quad$ $\quad$ $\quad$ 2.1.\  \  \textbf{For all $1\le j \le m$, Calculate $\gamma^j_{t}= \|g^j_t \|^2_{\h^j} $:}\\[0.1cm]
   \hspace{4cm} $\gamma^j_{t} =(1- \eta_t \lambda) ^2 \gamma^j_{t-1} +\left\langle K^j(x_t,x_t) \alpha_{t,t},\alpha_{t,t} \right\rangle_\y + 2(1- \eta_t \lambda)  \left\langle g^j_{t-1}(x_t), \alpha_{t,t} \right\rangle_\y $ \\[0.2cm] 
   $\quad$ $\quad$ $\quad$ 2.2.\  \  \textbf{For all $1\le j \le m$, Obtain $\delta^j_t$:}\\[0.1cm]
   \hspace{4cm} $\delta^j_t= \frac{\left( (\delta^j_{t-1})^2 \gamma^j_{t} \right)^{1/r+1} }{\left(\sum_j \left( (\delta^j_{t-1})^2 \gamma^j_{t} \right) ^{r/r+1} \right)^{1/r}} $ \\[0.2cm] 
   $\quad$ 3.\  \  \textbf{$f_t$ update phase:}\\[0.1cm]
   $\quad$ $\quad$ $\quad$ 3.1\  \  \textbf{Obtain $g^i_t$:}\\[0.1cm]
   \hspace{4cm} $g^j_t=\sum_{i=1}^{t} K^j(x_i,\cdot) \alpha_{i,t}$ \\[0.2cm] 
     $\quad$ $\quad$ $\quad$ 3.2\  \  \textbf{Obtain $f_t$:}\\[0.1cm]
   \hspace{4cm} $f_t=\sum_{j=1}^{n} \delta^j_t g^j_t$ \\[0.2cm] 
    \hline
  \end{tabular}
  \end{center}
  \end{table*}

In this section we describe a new algorithm called MONORMA, which extend ONORMA to the multiple operator-valued kernel (MovKL) framework \citep{kadri12}. The idea here is to learn in a sequential manner both the vector-valued function $f$ and a linear combination of operator-valued kernels. This allows MONORMA to learn the output structure rather than to pre-specify it using an operator-valued kernel which has to be chosen in advance as in ONORMA.

Let $K^1$,...,$K^m$ be $m$ different operator valued kernels  associated to RKHS $\h^1$,..,$\h^m \subset \y ^\x$. In batch setting, MovKL consists in learning a function  $\mathbf{f}=\sum_{i=1}^m \delta^i g^i \in \mathbf{E} = \h^1 + ... + \h^m$  with $g^i \in \h^i$ and
$(\delta^i)_i \!\in\! D_r\!=\!\left\lbrace (\omega^i)_{1 \le i \le m},\!\text{ such that }\! \omega^i\!>\!0 \text{ and } \sum_i (\omega^i)^r\!\le\! 1 \right\rbrace$, 
which minimizes  
$$\min_{(\delta^i_t) \in D} \min_{g^i_t \in \h^i} \frac{1}{t} \sum_{j=1}^t \l(\sum_i \delta^i_t g^i_t,x_j,y_j) + \sum_i \| g^i_t\|_{\h^i}.$$
%
%
To solve this minimization problem, the batch algorithm proposed by \citet{kadri12} alternatively optimize until convergence: 1) the problem with respect to $g^i_t$ when $\delta^i_t$ are fixed, and 2) the problem with respect to $\delta^i_t$ when $g^i_t$ are fixed.  Both steps are feasible. Step 1  is a classical learning algorithm with a fixed operator-valued kernel, and Step 2 admits the following closed form solution \cite{micchelli05}
\begin{equation}\label{d}
 \delta^i_t \leftarrow \frac{(\delta^i_t \|g^i_t\|_{\h^i})^\frac{2}{r+1}}{(\sum_i (\delta^i_t \|g^i_t\|_{\h^i})^\frac{2r}{r+1})^\frac{1}{r} }.
\end{equation}

It is important to note that the function $\mathbf{f}$ resulting from the aforementioned algorithm can be written $\mathbf{f}=\sum_j K(x_j,\cdot) \alpha_j $ where $K(\cdot,\cdot) = \sum_i d_i K_i(\cdot,\cdot) $ is the resulting operator-valued kernel.

The idea of our second algorithm, MONORMA,  is to combine the online operator-valued kernel learning algorithm ONORMA and the alternate optimization of MovKL algorithm. More precisely, we are looking for functions $f_t = \sum_i \delta^i_t  \sum_{j=0}^t K_i(x_j,\cdot) \alpha_{j,t}\in \mathbf{E}$. Note that $f_t$ can be rewritten as $f_t=\sum_i \delta^i_t g^i_t$ where  $g^i_t =  \sum_{j=0}^t K_i(x_j,\cdot) \alpha_{j,t}\in \h_ i$. At each iteration of MONORMA, we update successively  the $\alpha_{j,t}$ using the same rule as ONORMA,  and the $\delta^i_t$ using \eqref{d}.

\textbf{Description of MONORMA.}
For each training example $(x_t,y_t)$, MONORMA proceeds as follows: first, using the obtained value $\delta^i_{t-1}$ the algorithm determines $\alpha^i_t$ using a stochastic gradient descent step as in NORMA (see step 1.1 and 1.2 of MONORMA). Then, it computes $\gamma^i_t=\|g^i_t \|^2_{\h^i}$ using $\gamma^i_{t-1}$ and $\alpha^i_t$ (step 2.1) and deduces $\delta^i_{t}$ (step 2.2). Step 3 is optional and allows to obtain the functions $g^i_t$ and $f^t$  if the user want to see the evolution of these functions after each update.

It is important to note that in MONORMA,  $\|g^i_t\|$ is calculated without computing $g^i_t$ itself. This is motivated by the observation that the computational complexity of computing $\|g^i_t\|$  directly from the function $g^i_t$ has complexity $O(t^2 d^3)$ and thus is costly in an online setting (recall that $t$ denotes the number of inputs and $d$ denotes the dimension of $\y$). To avoid the need to do the computation of  $g^i_t$, one can use the following trick to save the time by calculating $\|g^i_t\|^2_{\h^i}$ from $\|g^i_{t-1}\|^2_{\h^i}$ and $\alpha_{t,t}$:
\begin{equation}\label{eq gamma}
\begin{aligned}
 \|&g^i_{t}\|^2_{\h^i} = \|(1- \eta_t \lambda) g^i_{t-1}+ K^i(x_{t}, \cdot)\alpha_{t,t} \|^2_{\h^i}  \\
&= 2(1- \eta_t \lambda) \left\langle g^i_{t-1},K^i(x_{t}, \cdot)\alpha_{t,t} \right\rangle_{\h^i} \\
&+(1- \eta_t \lambda)^2 \|g^i_{t-1}\|^2_{\h^i} +\left\langle K^i(x_{t}, \cdot)\alpha_{t,t},K^i(x_{t}, \cdot)\alpha_{t,t} \right\rangle_{\h^i}\\
&=2(1- \eta_t \lambda) \left\langle g^i_{t-1}(x_t),\alpha_{t,t} \right\rangle_{\y} +(1- \eta_t \lambda)^2 \|g^i_{t-1}\|^2_{\h^i}\\
& +\left\langle K^i(x_{t}, x_t)\alpha_{t,t},\alpha_{t,t} \right\rangle_{\y},
\end{aligned}
\end{equation}
where in the last transition we used the reproducing property. All of the terms that appear in \eqref{eq gamma} are easy to compute: $g^i_{t-1}(x_t)$ is already calculated during step 1.1 to obtain the gradient, and $\alpha_{t,t}$ is obtained from step 1. Hence, computing  $\|g^i_t\|^2$ from \eqref{eq gamma}  has complexity $O(d^2)$

The main advantages of ONORMA and MONORMA  their ability to approach the performance of their batch counterpart with modest memory and low computational cost. Indeed, they do not require the inversion of the operator-valued kernel matrix which is extremely costly. Moreover, in contrast to previous operator-valued kernel learning algorithms~\citep{Minh11, kadri12, Sindhwani13}, ONORMA and MONORMA can also be used to non separable operator-valued kernels. Finally, a major advantage of MONORMA is that interdependencies between output variables  are estimated instead of pre-specified as in ONORMA,  which means that the important output structure does not have to be known in advance.
%

\textbf{Complexity analysis.}
Here we consider a naive implementation of the MONORMA algorithm.
At iteration $t$, the $\alpha$ update step has complexity $O(t d)$. The $\delta$ update step has complexity $O(m d)$. Evaluating $g_j$ and $f$ has  complexity $O(t d^2 m)$. Hence the complexity of the iteration $t$ of \mbox{MONORMA} is $O(t d^2 m)$ and the complexity up to iteration $t$ is respectively $O(t^2 d^2 m)$.

\section{Experiments}\label{sect xp}

\begin{figure*}[!t]
\centering
\includegraphics[scale=0.4]     {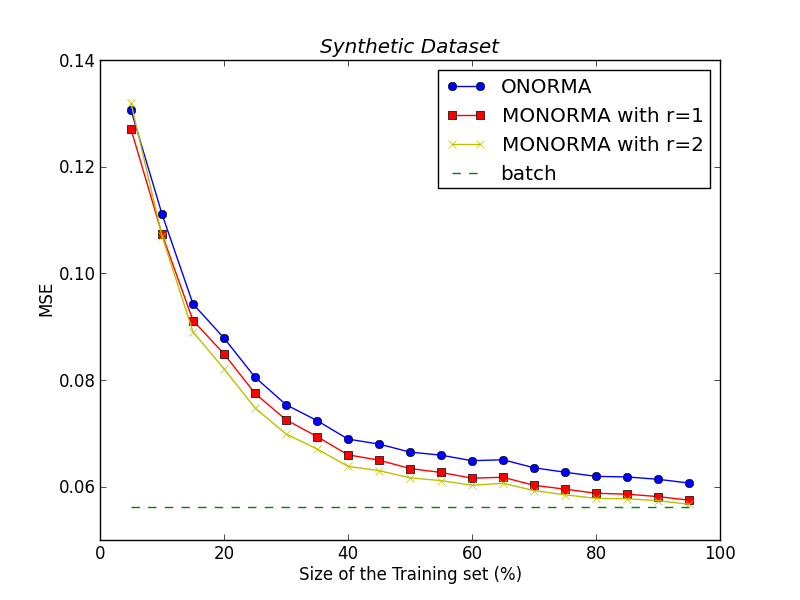}
\includegraphics[scale=0.4]     {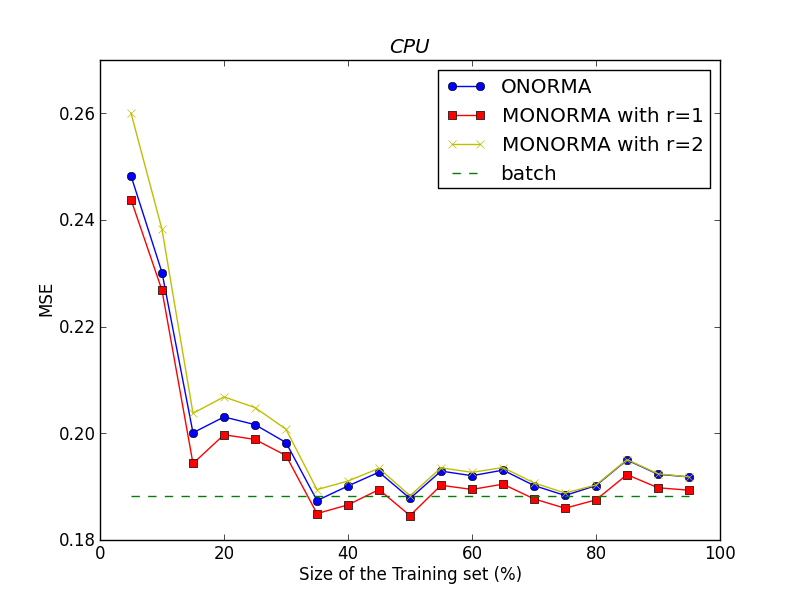}
\caption{Variation of MSE (Mean Square Error) of ONORMA (blue) and MONORMA (red and yellow) with the size of the dataset. The MSE of the batch algorithm (green) on the entire dataset is also reported. (left) Synthetic Dataset. (right) CPU Dataset.}
\label{fig_training}
\end{figure*}

In this section, we conduct experiments on synthetic and real-world datasets to evaluate 
the efficiency of the proposed algorithms, ONORMA and MONORMA. In the following, we set $\eta=1$, $\eta_t= 1/\sqrt{t}$, and $\lambda=0,01$ for ONORMA and MONORMA. We compare these two algorithms with their batch counterpart algorithm\footnote{The batch algorithm refers to the usual regularized least squares regression with operator-valued kernels (see, e.g., \citealt{kadri10}).}, while the regularization parameter  $\lambda$ is chosen with a five-fold cross validation. 

We use the following real-world dataset: Dermatology\footnote{Available at UCI repository \url{http://archive.ics.uci.edu/ml/datasets}.} (1030 instances, 9 attributes, $d=6$), CPU performance\footnote{Available at \url{http://www.dcc.fc.up.pt/~ltorgo/Regression}.} (8192 instances, 18 attributes, $d=4$), School data\footnote{see \citet{Argyriou2008}. } (15362 instances, 27 attributes, $d=2$). 
 Additionally, we also use a synthetic dataset (5000 instances, 20 attributes, $d=10$) described in \cite{Evgeniou05}. In this dataset, inputs~$(x_1, . . . , x_{20})$ are generated independently and uniformly over $\left[ 0, 1\right]$ and the different output are computed as follows. Let $\phi(x)=(x_1^2,x_4^2,x_1x_2,x_3x_5,x_2,x_4,1)$
and ($w^i$) denotes \textit{iid} copies of a $7$ dimensional Gaussian distribution with zero mean and covariance equal to Diag$(0.5; 0.25; 0.1; 0.05; 0.15; 0.1; 0.15)$.
 Then, the outputs of the different tasks are generated as $y^i=w^i \phi(x)$. We use this dataset with 500  instances and $d=4$, and  with 5000 instances and $d=10$.

For our experiments, we used two different operator-valued kernels: 
\begin{itemize}
\vspace{-0.4cm}
\itemsep=-0.1cm
 \item 
 a separable Gaussian kernel $K_\mu(x,x')=\exp(-\|x-x'\|_2^2 / \mu) \mathbf{J}$ where $\mu$ is the parameter of the kernel varying from $10^-3$ to $10^2$ and $\mathbf{J}$ denotes the $d \times d$ matrix with coefficient $\mathbf{J}_{i,j}$ equals to $1$ if $i=j$ and $1/10$ otherwise.
\item  
a non separable kernel $K_\mu(x,x')= \mu \left\langle x,x'\right\rangle \mathbf{1} + (1-\mu) \left\langle x,x'\right\rangle^2 \mathbf{I}$ where $\mu$ is the parameter of the kernel varying from $0$ to $1$ and $\mathbf{1}$ denotes the matrix of size $d \times d$ with all its elements equal to $1$ and $\mathbf{I}$ denotes the $d \times d$ identity matrix.
\end{itemize}

To measure the performance of the algorithms, we use the Mean Square Error (MSE) which refers to the mean cumulative error $ \frac{1}{\vert Z \vert} \sum\|f_{i-1}(x_i) - y_i\|_\y^2$ for ONORMA and MONORMA  and to the mean empirical error $ \frac{1}{\vert Z \vert} \sum\|f(x_i) - y_i\|_\y^2$ for the batch algorithm.

\begin{figure*}[!t]
\centering
\includegraphics[scale=0.4]     {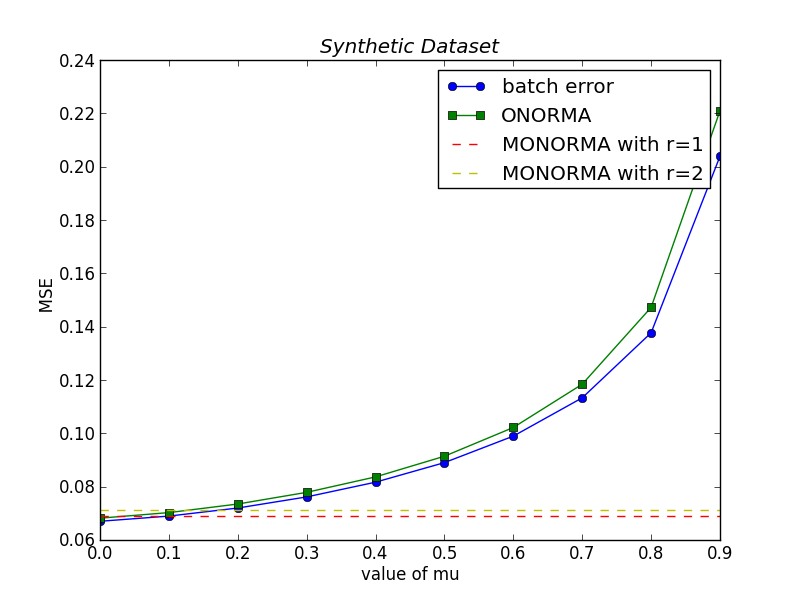}
\includegraphics[scale=0.33]     {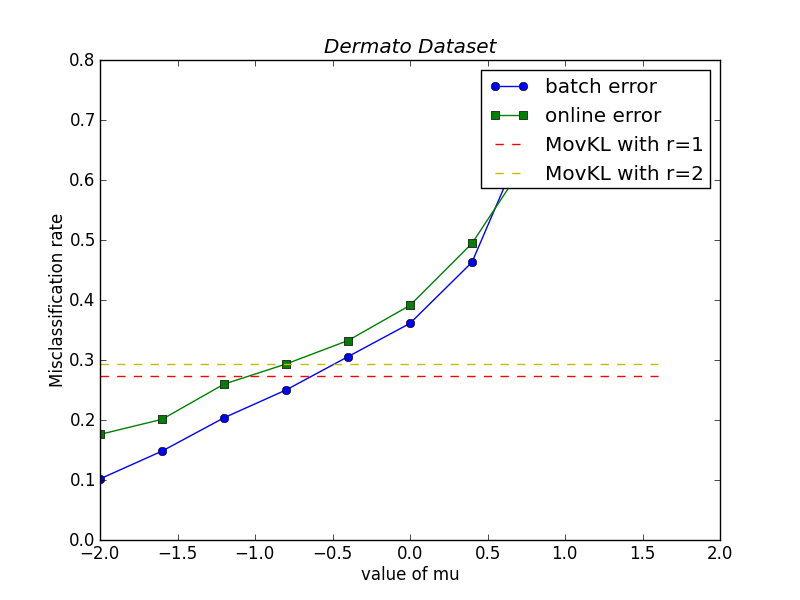} \\
\includegraphics[scale=0.4]     {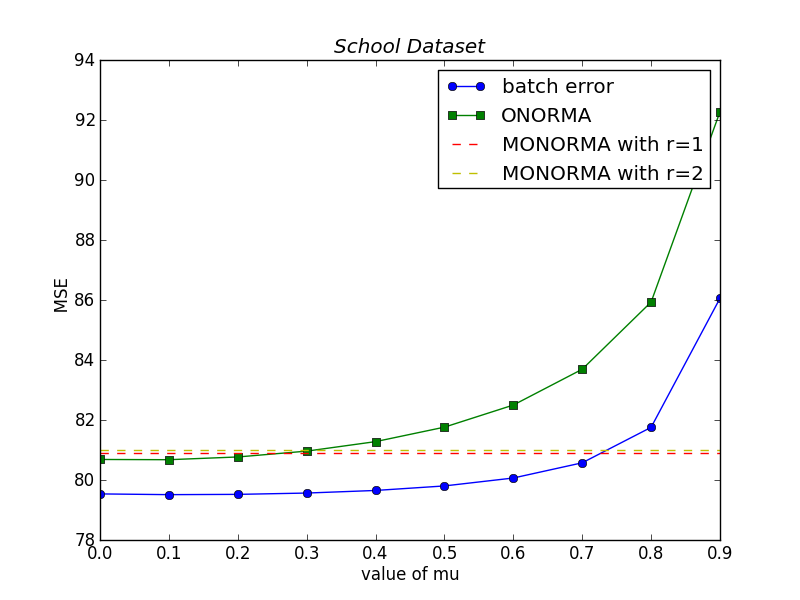}
\includegraphics[scale=0.4]     {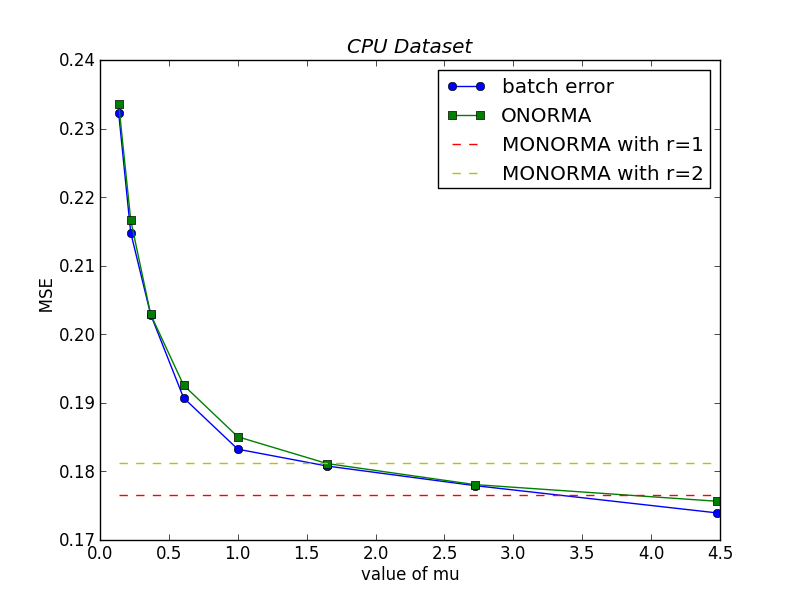}
\caption{ MSE of ONORMA (green) and the batch algorithm (blue) as a function of the parameter $\mu$. The MSE of MONORMA is also reported. (top left) Synthetic Dataset. (top right) Dermatology Dataset. (bottom left) CPU Dataset. (bottom right) School Dataset.}
\label{fig_kernel}
\end{figure*}

\textbf{Performance of ONORMA and MONORMA.}  In this experiment, the entire dataset $Z$ is randomly split in two parts of equal size, $Z_1$ for training and $Z_2$ for testing. 
Figure \ref{fig_training} shows the MSE results obtained with ONORMA and MONORMA on the synthetic and the CPU performance datasets when the size of the dataset increases, as well as the results obtained with the batch algorithm on the entire dataset. For ONROMA, we used the non separable kernel with parameter $\mu=0.2$ for the synthetic dataset and the Gaussian kernel with parameter $\mu=1$ for the CPU performance dataset. For MONORMA, we used the kernels $\left\langle x,x'\right\rangle \mathbf{1}$ and $\left\langle x,x'\right\rangle^2 \mathbf{I}$ for the synthetic datasets, and three Gaussian kernels with parameter $0.1$, $1$ and $10$ for the CPU performance dataset.
In Figure~\ref{fig_kernel}, we report the MSE (or the misclassification rate for the Dermatology dataset) of the batch algorithm and ONORMA when varying the kernel parameter $\mu$. The performance of MONORMA when learning a combination of operator-valued kernels with different parameters is also reported in Figure~\ref{fig_kernel}.
Our results show that ONORMA and MONORMA achieve a level of accuracy close to the standard batch algorithm, but  a significantly lower running time (see Table~\ref{table_time}). Moreover, by learning a combination of operator-valued kernels and then learning the output structure, MONORMA performs better than ONORMA.




\begin{table}
\small
\begin{center}

\caption{Running time in seconds of ONORMA, MONORMA and batch algorithms. }
\vspace{0.3cm}

\begin{tabular}{lcccc}

\hline
& Synthetic   & Dermatology & CPU\\
 \hline
Batch&162&11.89&2514\\
ONORMA&12&0.78&109\\
MONORMA&50&2.21&217\\
\hline
\end{tabular}
 \label{table_time}
\end{center}

\end{table}

\section{Conclusion}

In this paper we presented two algorithms: ONORMA is an online learning algorithm that extends NORMA to operator-valued kernel framework and requires the knowledge of the output structure. MONORMA does not require this knowledge since it is capable to find  the output structure by learning sequentially a linear combination of operators-valued kernels. We reported a cumulative error bound for ONORMA that holds both for classification and regression. We provided experiments on the performance of the proposed algorithms that demonstrate variable effectiveness. Possible future research directions include 1) deriving cumulative error bound for MONORMA, and 2) extending MONORMA to structured output learning. 

\bibliographystyle{plainnat}
\bibliography{biblioinfo_online}

\end{document}